\documentclass{IEEE_lsens}
\usepackage{textcomp}
\usepackage{hyperref}
\usepackage{url}
\usepackage[linesnumbered]{algorithm2e}
\usepackage{graphicx}
\usepackage{caption}
\usepackage{subcaption}
\usepackage{multirow}
\usepackage[normalem]{ulem}
\useunder{\uline}{\ul}{}
\usepackage[compact]{titlesec}

\usepackage[noadjust]{cite}
\ifCLASSINFOpdf
\else
\fi
\usepackage[T1]{fontenc} % optional enhanced font encoding
\usepackage{amsmath}
\usepackage[cmintegrals]{newtxmath}
\usepackage{bm}
\usepackage{array}
\usepackage{url}
\ifCLASSINFOpdf
\else
\fi
\providecommand{\hypersetup}[1]{\relax}

\begin{document}

% The paper headers
% \markboth{Vol.~1, No.~3, July~2017}{0000000}

% article subject line 
% \IEEELSENSarticlesubject{Sensor Applications}

% paper title
% Titles are generally capitalized except for words such as a, an, and, as,
% at, but, by, for, in, nor, of, on, or, the, to and up, which are usually
% not capitalized unless they are the first or last word of the title.
% Linebreaks \\ can be used within to get better formatting as desired.
% Do not put math or special symbols in the title.
%
\title{Estimation of Clinical Workload and Patient Activity using Deep Learning and Optical Flow}

\author{
\IEEEauthorblockN{Thanh Nguyen-Duc\IEEEauthorrefmark{1}$^,$\IEEEauthorieeemembermark{1}, Peter Y Chan\IEEEauthorrefmark{1}$^,$\IEEEauthorrefmark{2}$^,$\IEEEauthorieeemembermark{1}, Andrew Tay\IEEEauthorrefmark{2},
David Chen\IEEEauthorrefmark{2}, John Tan Nguyen\IEEEauthorrefmark{1}, Jessica Lyall\IEEEauthorrefmark{2} and Maria De Freitas\IEEEauthorrefmark{2}}% <-this % stops a space
% \author{\IEEEauthorblockN{Michael~Shell\IEEEauthorrefmark{1}\IEEEauthorieeemembermark{1}, John~Doe\IEEEauthorrefmark{2},
% and~Jane~Doe\IEEEauthorrefmark{2}\IEEEauthorieeemembermark{2}}% <-this % stops a space
\IEEEauthorblockA{\IEEEauthorrefmark{1}Monash University, Melbourne, Victoria, Australia \\
\IEEEauthorrefmark{2}Eastern Health, Melbourne, Victoria, Australia}
\thanks{\IEEEauthorieeemembermark{1}Equal contributions. }}%
%Emails: thanh.nguyen4@monash.edu, peter.chan@easternhealth.org.au
\IEEEtitleabstractindextext{%
\begin{abstract}
Contactless monitoring using thermal imaging has become increasingly proposed to monitor patient deterioration in hospital, most recently to detect fevers and infections during the COVID-19 pandemic. In this letter, we propose a novel method to estimate patient motion and observe clinical workload using a similar technical setup but combined with open source object detection algorithms (YOLOv4) and optical flow. Patient motion estimation was used to approximate patient agitation and sedation, while worker motion was used as a surrogate for caregiver workload. Performance was illustrated by comparing over 32000 frames from videos of patients recorded in an Intensive Care Unit, to clinical agitation scores recorded by clinical workers.
\end{abstract}

}

% make the title area
\maketitle

%%%%%%%%%%%%%%%%%%%%%%%%%%%%%%%%%%%%%%%%%%
\section{Introduction}
 High caregiver workload has been implicated in an increased risk of burnout, potentially reducing workforce capacity and worsening patient outcomes \cite{portoghese2014burnout}. This has been made worse through higher demands associated with the COVID-19 pandemic \cite{shoja2020covid}. While many semi quantitative metrics have been suggested as ways of monitoring workload \cite{bruyneel2019measuring}, the use of computer vision to directly monitor and quantify caregiver-patient interaction and task performance could potentially provide an objective measurement of caregiver workload. While computer vision has been used previously to monitor wellbeing in animal studies \cite{lencioni2021pain}, to our knowledge it has never been used to monitor staffing workload.

Multiple studies have used deep learning to monitor patients. For example, Anis et al. suggested the use of cameras to monitor facial expressions in patients for main and discomfort~\cite{davoudi2019intelligent}. Bedridden patient positions have been observed by using micro-doppler signatures~\cite{nazaroff2021tracking}. Respiratory rate and other vital signs have been monitored using thermal cameras coupled with deep learning by Lyra et al.  \cite{lyra2021deep}. We propose the use of dense optical flow to estimate both patient motion, which can be correlated to a clinical sedation and agitation score, as well as caregiver motion and their proximity to the patient, which is used as a corollary to workload. 

In this letter, we propose a method to automatically observe patient motion and physical interactions with healthcare workers using a single low resolution thermal camera that would maintain privacy. Our contributions include a) the use of a single low cost thermal camera sensor trained to recognize and monitor patient motion, caregiver workload and worker-patient interaction ; b) A real world illustration of this method's performance comparing existing clinical metrics with recorded videos.

%TODO Peter \cite{everingham2010pascal} \& Thanh Fig\ref{fig:overview} 
%%%%%%%%%%%%%%%%%%%%%%%%%%%%%%%%%%%%%%%%%%
\section{Method}

\begin{figure}[t]
\includegraphics[width=1\columnwidth]{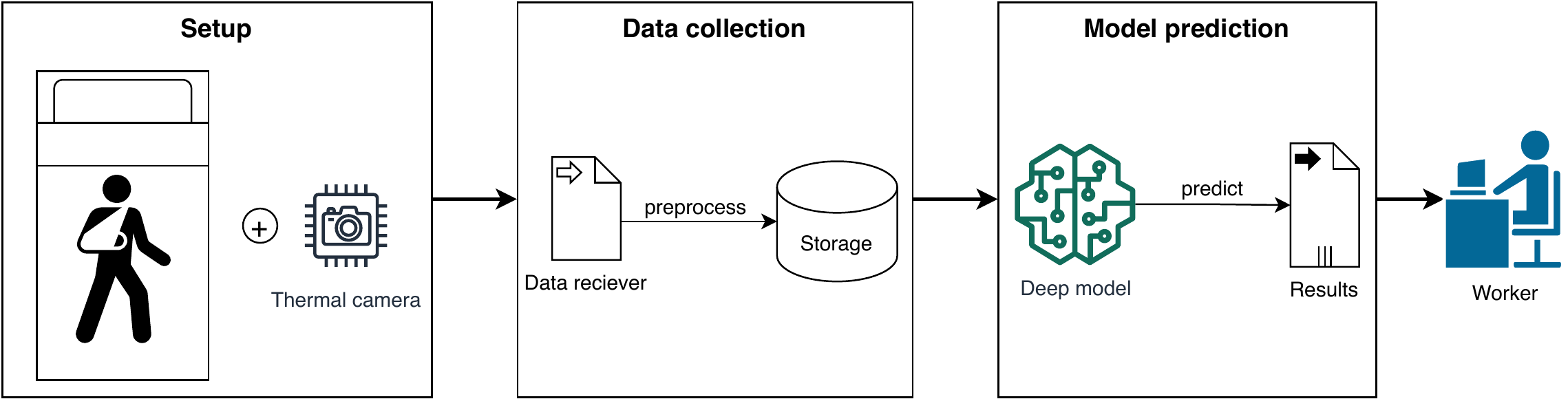}
\caption{Framework overview including three steps: thermal camera setup, data collection and model prediction. The healthcare worker can easily observer the patient via result report output from the framework.}
\label{fig:overview}
\end{figure}

An overview of the proposed method is displayed in Fig.~\ref{fig:overview}. A thermal camera was set up to continuously capture video from a clinical bedspace in a metropolitan Intensive Care Unit (ICU). Video frames were then sent to a server for preprocessing and storage.
A deep learning model (e.g., YOLOv4~\cite{bochkovskiy2020yolov4}) was then employed to detect the bounding box locations of the patient and healthcare workers in the frame. The system then used the boxes to record interaction time between the healthcare workers and the patient, calculated by leveraging the overlap of these boxes. Specifically, interactions were counted when the box overlapping between patient and worker was larger than a defined threshold. Motion estimation of the patient was then simultaneously estimated by using dense optical flow~\cite{farneback2003two}. 

\subsection{Thermal camera}
%, a field of view (FOV) 28.7º (H) × 21.7º (V)—35.3º (diagonal), a sensitivity of < 50mK@F/1 and < 80mK@F/1.3, and a frame rate of 4 Hz.
Thermographic footage was obtained with the Thermal Experts TE-Q1 narrow-angle camera at 384 $\times$ 288 resolution Cameras were connected to a Raspberry Pi 2 Model B (Cambridge, UK), streaming raw unprocessed files as Numpy arrays of temperature measurements to a portable storage device.

\subsection{Data collection}
Data was collected as part of a prospective observational single centre study in the ICU department of a major metropolitan hospital (Box Hill Hospital) in Melbourne between March 2019 and May 2020. Ethics approval was obtained from the Eastern Health Human Research Ethics Committee (LRR 033/2017) and conducted in accordance with the Declaration of Helsinki.  Patients admitted into an Intensive Care Unit room with the thermal imaging camera installed were eligible. Informed consent was obtained from patients allowing thermal imaging recording, the usage of thermal data and access to their clinical record. Patients were recorded continuously throughout their stay.

The thermal imaging cameras were mounted on the ceiling of an ICU room 2.2m away from the patients at a 30 degree angle. The ICU room measures 5.2m $\times$ 4.2m $\times$ 27m. Room temperature in the ICU room is controlled to 22 degrees Celsius. Thermal images were 384$\times$288 resolution, with each "pixel" containing a temperature reading in Celsius. Using OpenCV in python, the Numpy arrays were converted to JPEG with appropriate brightness and contrasts adjustments applied to yield a black and white image.

\subsection{Patient and worker observation model}

\begin{figure}[t]
\includegraphics[width=1\columnwidth]{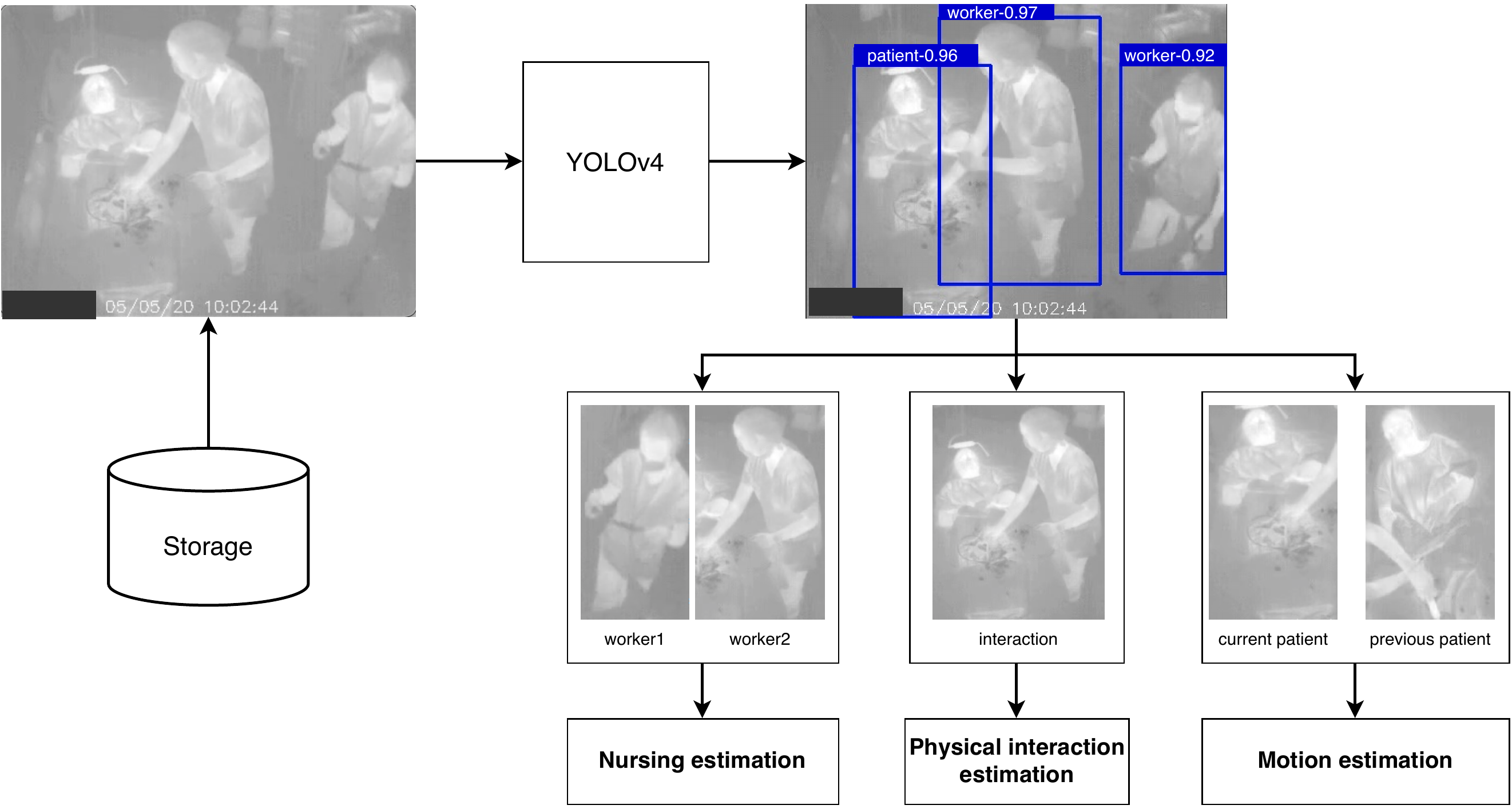}
\caption{Patient observation model pipeline takes video frames. YOLOv4 performs object detection task. Patient and worker locations are then process to do nursing and motion estimation.}
\label{fig:model}
\end{figure}

A pipe line of the model is on Fig.~\ref{fig:model}. 

The backbone module of the method was to leverage a trainable object detection model (e.g., YOLOv4~\cite{bochkovskiy2020dyolov4}) that determined the object locations on the video frame where certain patients and workers are present, as well as classifying those objects. In Fig.~\ref{fig:model}, video frames are first inputted into YOLOv4, which detected the locations of workers and patients. Based on the detection outputs, we further improved performance of our patient observation estimations by eliminating extraneous information including the background and other room objects. 

\subsubsection{Object detection using YOLOv4}

You Only Look Once~\cite{redmon2016you} (YOLO) is a deep neural network model used for real-time object detection. It significantly outperforms others object detection (e.g., R-CNN~\cite{girshick2014rich}) in term of running time, which was necessary in our use case. YOLO has been further improved by adding batch normalization \cite{ioffe2015batch} and feature pyramid networks  such as in YOLOv2~\cite{redmon2017yolo9000} and YOLOv3~\cite{redmon2018yolov3}. In this paper, we exploited YOLOv4~\cite{bochkovskiy2020yolov4}, which applied CIoU loss~\cite{zheng2020distance} to achieve optimal speed and accuracy in object detection tasks. 

Object detection in our framework can be understood as a function $f$. This function takes an video frame input image ($I$), then generate bounding boxes and its class prediction as shown in Eq.~\ref{eq:f}.

\begin{equation}
     f(I) =  \left\{
\begin{array}{ll}
      patient\_1 & bounding\_box\_location, \\
      worker\_1 & bounding\_box\_location, \\
      ... & ... \\
      patient\_n & bounding\_box\_location, \\
      worker\_m & bounding\_box\_location. \\
\end{array} 
\right.
\label{eq:f}
\end{equation}
where $n$, $m$ and $bounding\_box\_location$ is number of patients, number of workers, location of a detected object on input image respectively.

YOLOv4~\cite{bochkovskiy2020yolov4} was trained and optimized to a 0.91 mAP performance (see  the Experiments section for more details). It was then deployed to take an input video frame every second. It then separated locations of patients and workers with its class probabilities for next steps, shown in Fig.~\ref{fig:model}.

\subsubsection{Nursing time and physical interaction estimation}

\textbf{Nursing time} was defined as the total time that the patient admitted to the ICU was looked after by healthcare workers. Because the YOLOv4 generated predictions in a short interval time (e.g., one second in our framework), we were able to simply count the number of workers that appear in front of the thermal camera shown in Fig.~\ref{fig:model} to quantify. Nursing time in seconds was calculated in Eq.~\ref{eq:nurse}.
\begin{equation}
    nursing\_time = \sum_t m_t,
    \label{eq:nurse}
\end{equation}
where $t$ is one second.

\textbf{Physical interaction (PI)} between patient and worker was calculated to approximate caregiver workload. In our framework, we defined a physical interaction as when the union of worker and patient bounding boxes divided by patient bounding box was bigger than a threshold $\tau$ as shown in Eq.~\ref{eq:phy}. $\tau$ was set at 0.1. Similar to nursing time, we counted the number of physical interactions for each output prediction from YOLOv4~\cite{bochkovskiy2020yolov4} that was equivalent to one second in real-time.
\begin{equation}
    PI =  \left\{
    \begin{array}{ll}
          1 & \frac{patient\_bounding\_box ~\cup~ worker\_bounding\_box}{patient\_bounding\_box} \geq \tau, \\
          0 & otherwise. \\
    \end{array} 
    \right.
    \label{eq:phy}
\end{equation}

\subsubsection{Patient motion estimation} 
Patient motion was defined as a scalar number which indicated movement intensity of patient at time $t$ compared to patient's location from the previous frame at $t-dt$ by interval time $dt$ in the video.  Dense optical flow was used to estimate movement of objects between the two frames. The bounding box around the patient was denoted as an image  $I_{patient}(x, y, t-dt)$ with the $x, y$ pixel location in two dimensions at time $t-dt$. This image moves by distance $dx, dy$ in the next frame at time $t$. We can form:
\begin{equation}
    I_{patient}(x, y, t-dt) = I_{patient}(x + dx, y + dy, t)
\end{equation}

\begin{figure}[t]
 	\centering	
 	\begin{subfigure}{0.49\linewidth}
 		\centering
 		\includegraphics[width=1\linewidth]{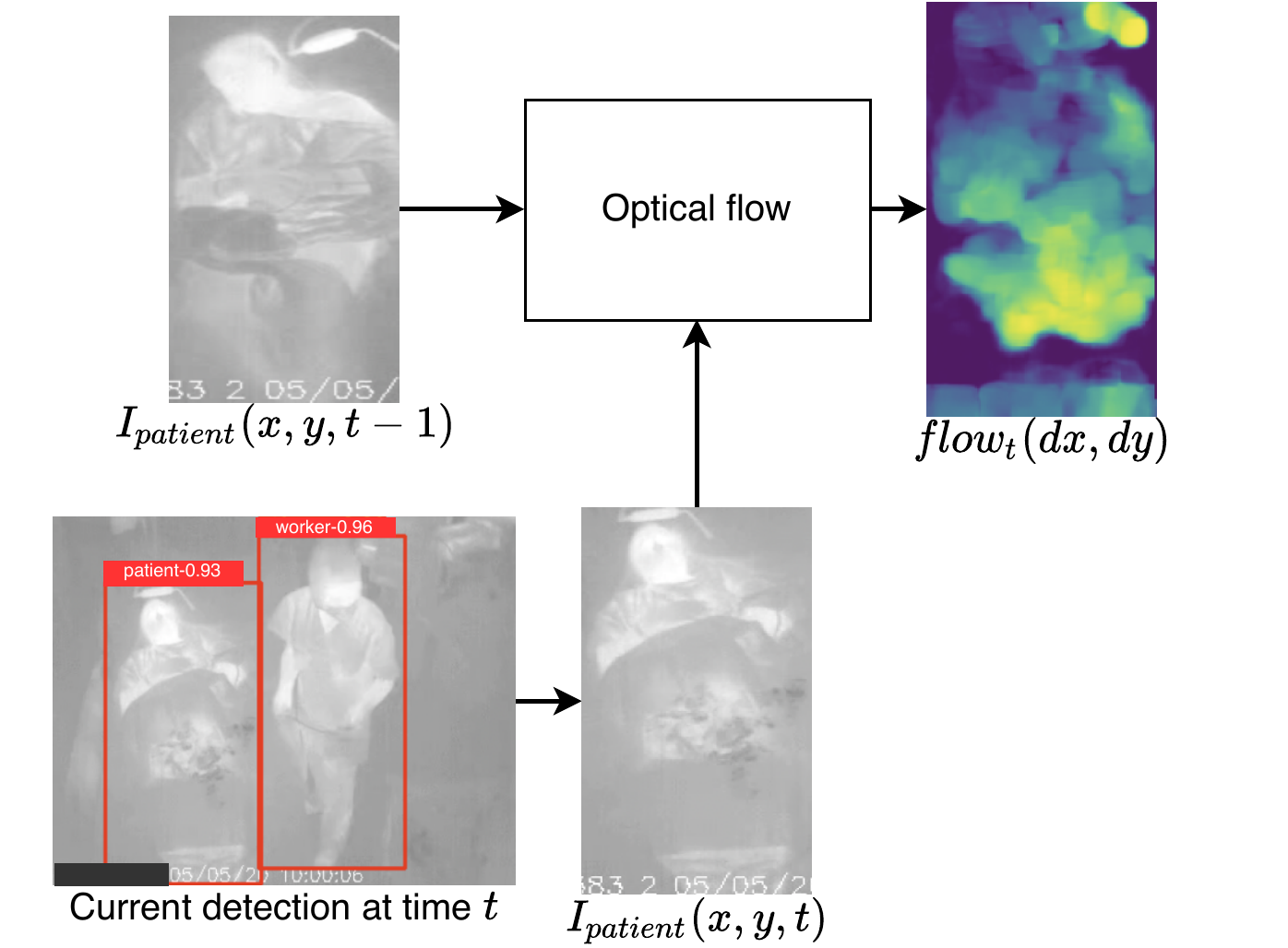}
		\caption{Optical flow when there is no physical interaction.}
		\label{fig:ot1}
 	\end{subfigure}
 	\hfill
 	\begin{subfigure}{0.49\linewidth}
 		\centering
 		\includegraphics[width=1\linewidth]{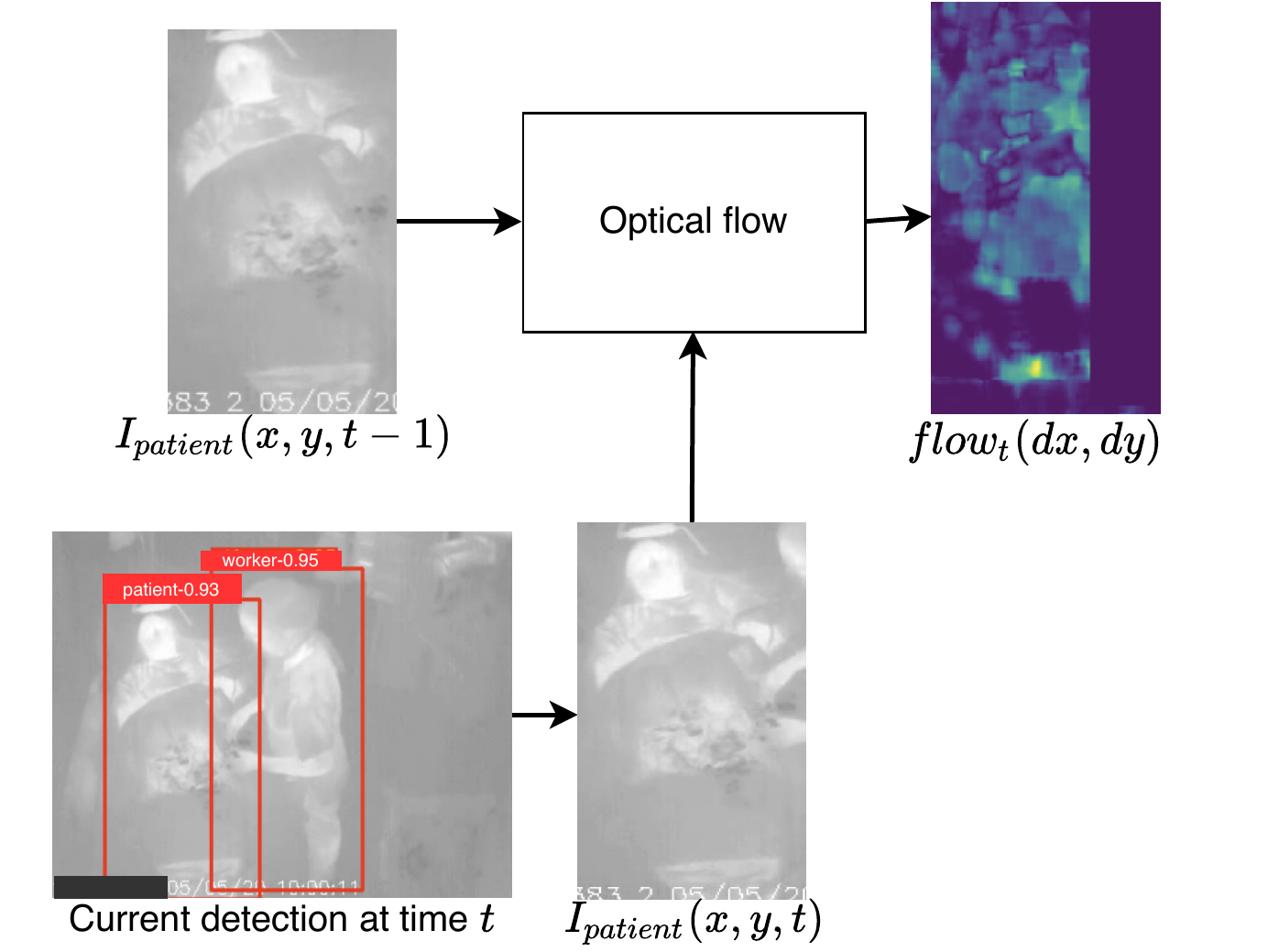}
		\caption{Refined optical flow when there is a physical interaction.}
	    \label{fig:ot2}
 	\end{subfigure}
 	\caption{Optical flow illustration in two different scenarios.}
 	\label{fig:optical}
 \end{figure}

Optical flow ~\cite{farneback2003two} was used to estimate movement vectors $flow(dx, dy)$ at every ($x,y$) pixel location shown in Fig.~\ref{fig:ot1}. This was further refined $flow(dx, dy)$ by subtracting physical interactions between the worker and patient shown in Fig.~\ref{fig:ot2}. Motion at time $t$ was computed by the $mean$ and the standard deviation $std$ of the flow magnitude $|| flow ||$. Motion is then relaxed by adding the previous motion as shown in Eq.~\ref{eq:motion}. In our framework, $\alpha$ = 0.7 and $dt$ = 1 second.

\begin{equation}
    motion_t = \alpha(mean(|| flow ||) + std(|| flow ||)) + (1-\alpha)motion_{t-1}
    \label{eq:motion}
\end{equation}

\section{Experiments}
In this section, the performance of our method is illustrated using various scenarios and metrics. YOLOv4 was individually evaluated using mean Average Precision (mAP) on 1000 human labeled test images. We then compared nursing time and physical interaction to ground-truth data (e.g., over 32k manually annotated video frames). We also show the correlation between our motion estimation and the Riker sedation-agitation score recorded by healthcare professionals during the patient stay.

\subsection{YOLOv4 evaluation}
\textbf{Setup.}
Optimal performance in our framework was achieved by training and evaluating YOLOv4 to maximize object detection ~\cite{bochkovskiy2020yolov4}. A dataset was generated by randomly selecting 5000 images from arbitrary videos in the storage. Bounding box locations of patients and workers were then manually annotated using the Computer Vision Annotation Tool\footnote{https://github.com/openvinotoolkit/cvat}. This dataset was then split to two 4000 images for training and 100 images for testing. YOLOv4 was implemented using Tensorflow2\footnote{https://github.com/pythonlessons/TensorFlow-2.x-YOLO}. YOLOv4 was trained using DarkNet53~\cite{redmon2018yolov3} feature extractor with Adam optimizer~\cite{kingma2014adam} on a Titan V GPU.

\textbf{Metric.} We evaluated performance of the object detection by using a well-known metric called mean Average Precision (mAP)\cite{everingham2010pascal}.
mAP calculates mean class precisions by comparing the ground-truth bounding box to the detected box and its class prediction given a intersection over union (IoU) threshold. Specifically, given a threshold is equal to 0.5, an object is correctly predicted when classification is the same to class label and its detected box overlaps more than 50\% (IoU=0.5) to the corresponding ground-truth. The higher mAP score (range from 0 to 1), the better performance is in the YOLOv4 detection. 

\textbf{Result.} mAP was calculated with different IoU thresholds to test the YOLOv4. In Table~\ref{table:map}, mAPs of more than 0.99 and 0.94 were achieved at 0.5 and 0.7 thresholds for patient and worker detection respectively. Overall mAP was approximately 0.91 for both patient and worker detection.

\begin{table}[t]
\caption{Mean Average Precision (mAP) on testing images.}
\label{table:map}
\centering
\begin{tabular}{|c|c|c|c} 
\cline{1-3}
        & Patient & Worker                                    &                                                                          \\ 
\cline{1-3}
mAP@0.5  & 0.99   & 0.95                                     &                                                                          \\ 
\cline{1-3}
mAP@0.7  & 0.99   & 0.95                                     &                                                                          \\ 
\cline{1-3}
mAP@0.9  & 0.87   & 0.70                                      &                                                                          \\ 
\hline
\textbf{Average} & \textbf{0.95}    & \textbf{0.87} & \multicolumn{1}{l|}{\textbf{0.91}}  \\
\hline
\end{tabular}

\end{table}

\subsection{Nursing time and physical interaction evaluation}
\textbf{Setup.} Caregiver time and physical interaction estimation was derived from three videos from three different patients, randomly sampled from the storage archive. Each of the videos was three hours in length and recorded at different times of the day (video 1 from 7AM-11AM, video 2 from 1PM-4PM, video 3 from 10AM-1PM). Nursing time and physical interactions were manually labelled every second. 

\textbf{Result.} The accuracy between the model prediction and human labelling between number of healthcare workers and physical interactions is shown in Table~\ref{table:acc}. In video 2, our framework achieved 0.98 and 0.979 accuracy on worker and physical interaction respectively. The average accuracy of the three videos was 0.956 and 0.98 for worker and interaction counting respectively.

The total nursing time derived from the model compared to human labelling is shown in Table~\ref{table:nurse}. The proposed framework estimated nursing time with an error of approximately 5 minutes. 

\begin{table}[t]
\caption{Accuracy of healthcare worker and physical interaction counting.}
\label{table:acc}
\centering
\begin{tabular}{|c|c|c|}
\hline
          & Worker counting & Physical interaction counting \\ \hline
Video 1 & 0.91            & 0.98                          \\ \hline
Video 2 & 0.98            & 0.979                         \\ \hline
Video 3 & 0.978           & 0.982                         \\ \hline
\textbf{Average}   & \textbf{0.956}           & \textbf{0.98}                          \\ \hline
\end{tabular}

\end{table}
\begin{table}[t]
\caption{Nusring time comparison.}
\label{table:nurse}
\centering
\begin{tabular}{|c|c|c|c|} 
\cline{1-4}
        & Predicted nursing time & Label nursing time & Error time \\ 
\cline{1-4}
Video 1 & 57m25s                        & 52m10s & 5m15s                                                              \\ 
\cline{1-4}
Video 2 & 1h28m13s                      & 1h30m23s    & 2m10s                                                         \\ 
\cline{1-4}
Video 3 & 1h00m21s                      & 1h05m29s      & 5m29s                                                         \\
\cline{1-4}
\end{tabular}

\end{table}

The Table~\ref{table:inter} shows a comparison between the predicted interaction time and manually labelled interaction time. The average error time between the model prediction and human labelling was less than 1 minute.
\begin{table}[t]
\caption{Physical interaction time comparison.}
\label{table:inter}
\centering
\begin{tabular}{|c|c|c|c|} 
\cline{1-4}
        & Predicted interaction time & Label interaction time & Error time \\ 
\cline{1-4}
Video 1 & 13m38s                                        & 14m10s      & 32s                                                           \\ 
\cline{1-4}
Video 2 & 38m12s                                        & 39m20s  &   1m8s                                                        \\ 
\cline{1-4}
Video 3 & 33m33s                                        & 34m48s  &  1m15s                                                             \\
\cline{1-4}
\end{tabular}

\end{table}

\begin{figure}[t]
 	\centering	
 	\begin{subfigure}{0.49\linewidth}
 		\centering
 		\includegraphics[width=1\linewidth]{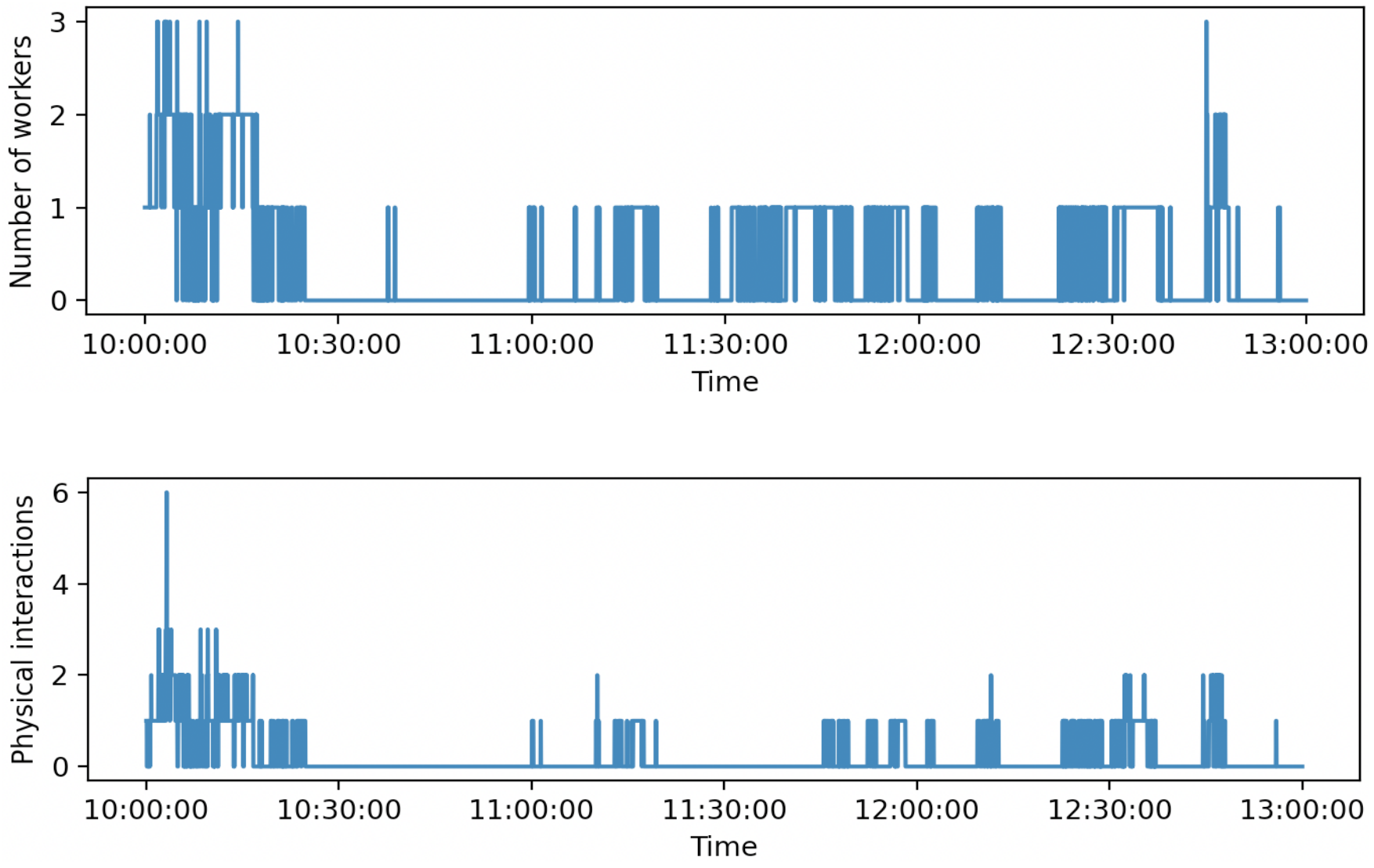}
		\caption{Prediction}
		\label{fig:pred}
 	\end{subfigure}
 	\hfill
 	\begin{subfigure}{0.49\linewidth}
 		\centering
 		\includegraphics[width=1\linewidth]{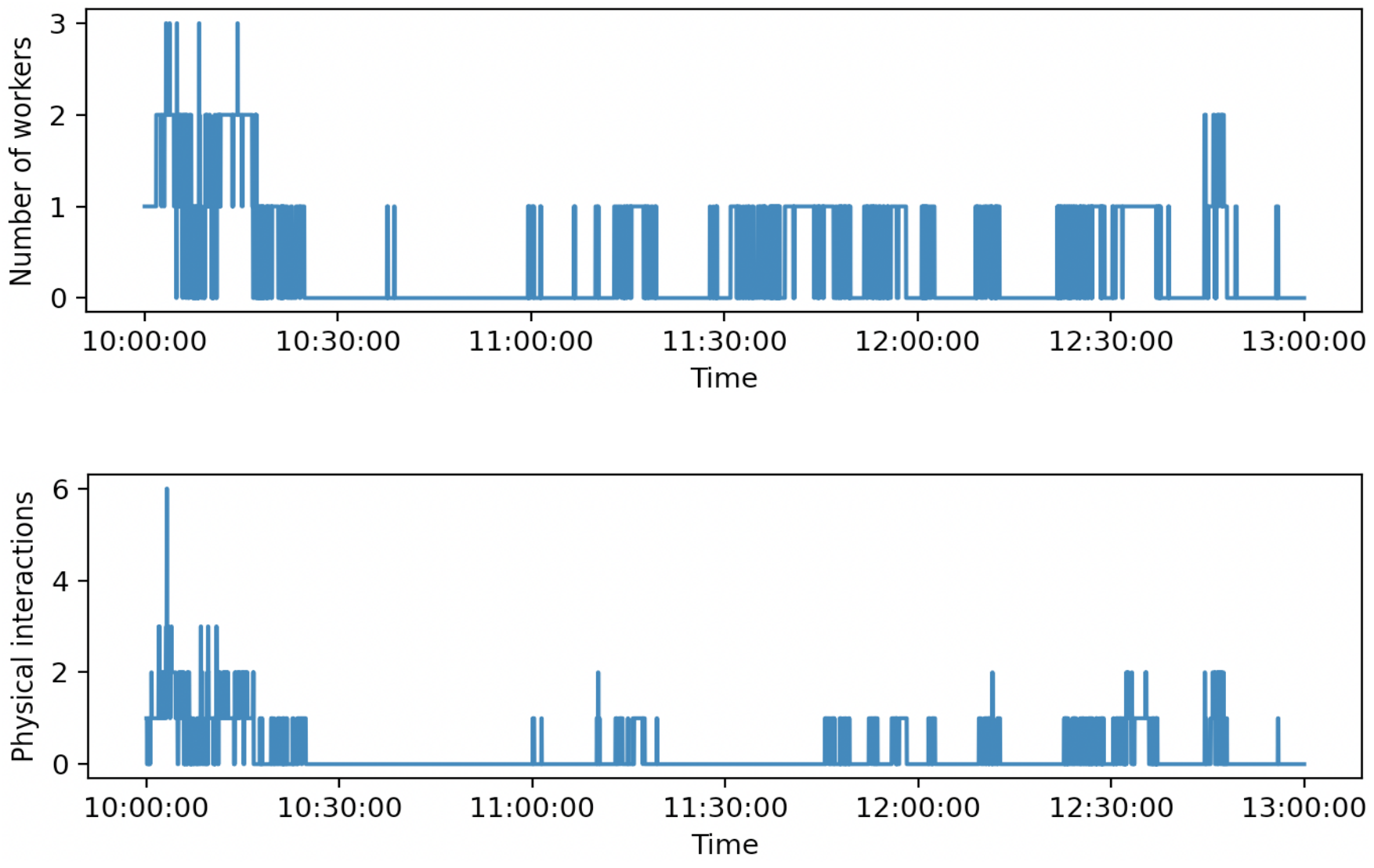}
		\caption{Ground-truth.}
	    \label{fig:gt}
 	\end{subfigure}
 	\caption{Comparison between model's prediction and ground-truth over three hours. First row: number of workers Second row: number of physical interactions.}
 		\label{fig:seq}
 \end{figure}

Video 3 was used to plot the number of workers and physical interactions for each second (Fig.~\ref{fig:seq}
). Fig~\ref{fig:pred} illustrates the same pattern compared to the ground-truth in Fig~\ref{fig:gt}.

\subsection{Motion estimation evaluation}
\begin{figure}[!t]
\centering	
\includegraphics[width=0.7\columnwidth]{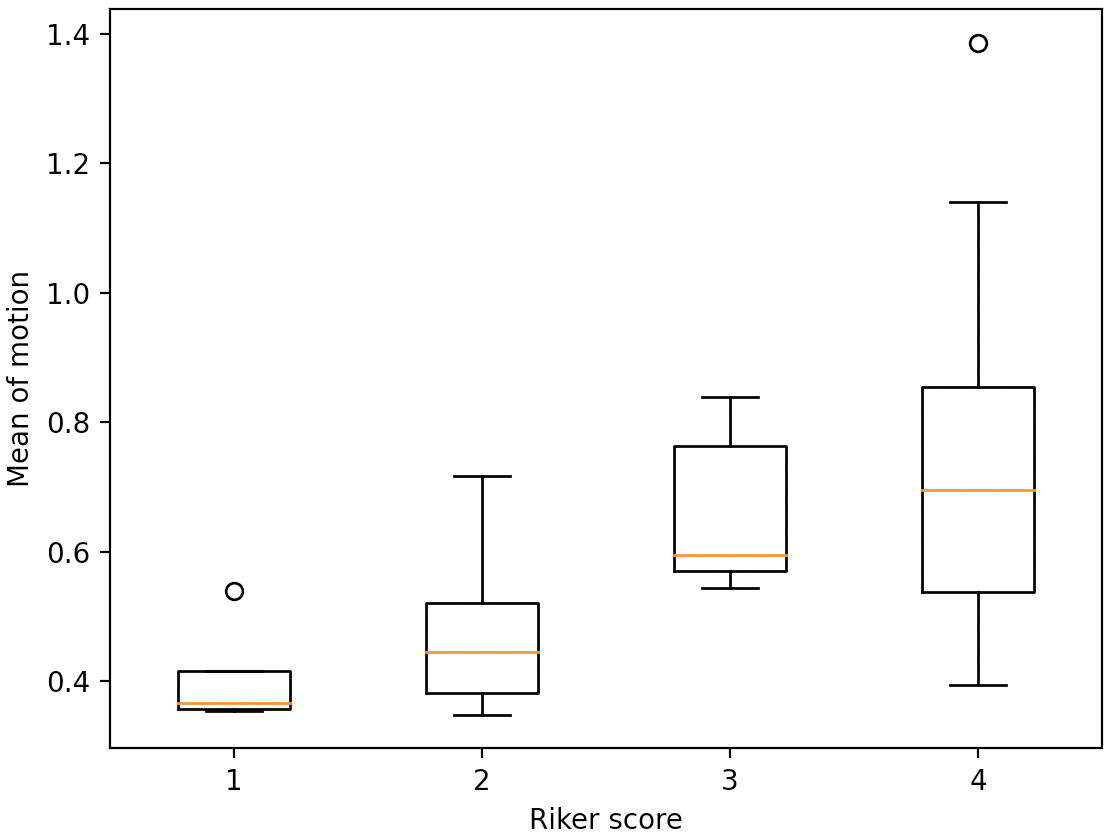}
\caption{Mean of predicted motion at different Riker scores.}
\label{fig:box}
\end{figure}
\begin{figure}[!t]
\centering	
\includegraphics[width=1\columnwidth]{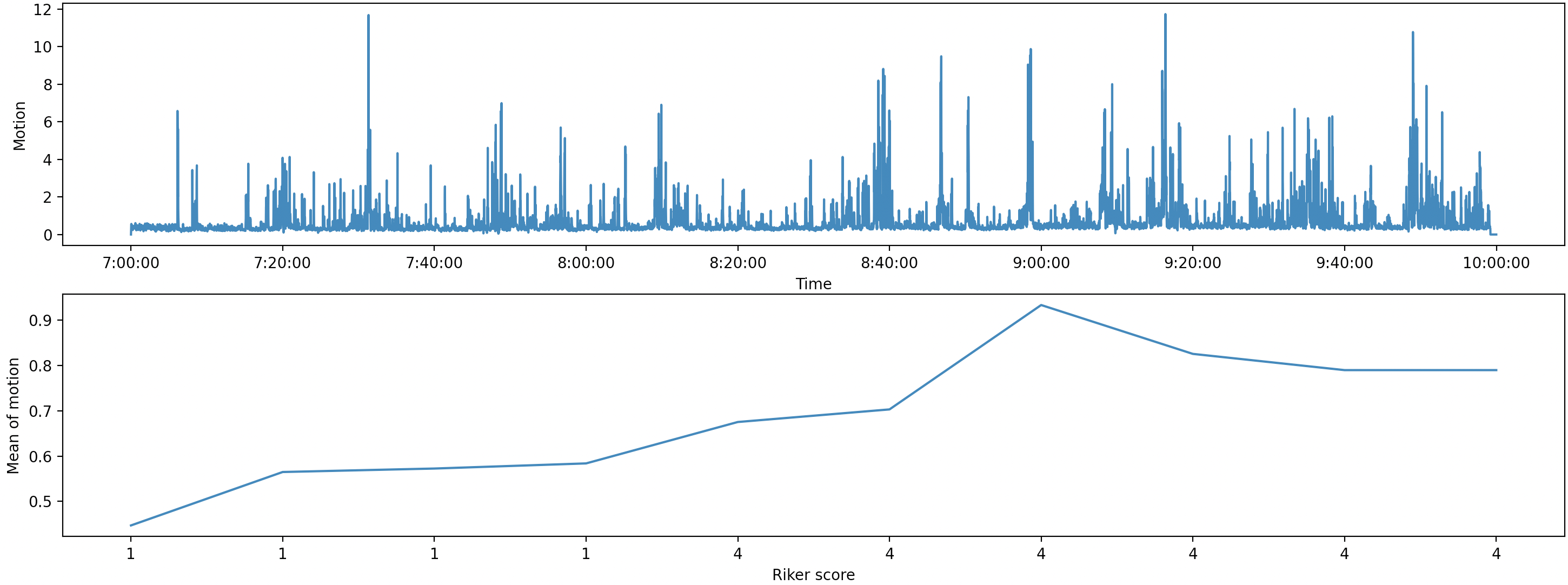}
\caption{First row: Motion over time. Second row: Correlation between mean of motion and Riker score.}
\label{fig:motion}
\end{figure}

\textbf{Setup.} Four videos with varying severities of agitated patient were selected. Each video also had instances where the same patient demonstrated multiple severities of agitation. The performance of our method was then evaluated by aligning the predicted motion and the Riker score recorded by nurses.

\textbf{Metric.} The Riker sedation agitation score is a commonly semiquantitative metric recorded by bedside caregivers and used to measure patient agitation range from 1 to 7, where 1 is an unrousable patient, and 7 means the patient is severely agitated.
 
\textbf{Result.}
Fig.~\ref{fig:box} demonstrates the median of mean motion, with the bottom and top edges of the box indicating 25th and 75th percentiles. The predicted mean motion demonstrates aligmment to the Riker score. The same result is illustrated in Fig.~\ref{fig:motion}.

\section{Conclusion}
By utilizing object detection and analyzing motion, the amount of time caregivers spend interacting with patients could be quantified and potentially allow healthcare managers can to efficiently roster staff and minimize overwork, potentially reducing staff burnout. Further, this same technique could be employed to monitorthe overall comfort and progress of patients.

In this letter, we propose a novel method to monitor both patient motion and caregiver workload using a single thermal camera. Our method demonstrates accurate detection of both patients and caregivers as well as the ability to estimate motion, caregiver time and physical interaction. We believe this method could potentially help better quantify caregiver workload and patient agitation in a healthcare setting without compromising patient and staff privacy. Future directions could potentially further improve method performance by fine-tuning YOLOv4 object detection and dense optical flow.

% use section* for acknowledgment
%\section*{Acknowledgment}
% addcontentsline needed when using bookmark hyperlinking, etc.
%\addcontentsline{toc}{section}{Acknowledgment}
% enable scriptsize
%\scriptsize
%This work was supported by the IEEE. The authors would
%like to thank ... Note that the Acknowledgment section of
%\textit{IEEE Sensors Letters} is rendered in scriptsize.

% put at least one blank line to end the scriptsize paragraph and
% then revert back to normalsize.
\normalsize

% Last page column equalization
%
% IEEE Sensors Letters does balance the columns on the last page.
% Can use:
% \IEEEtriggeratref{8}
% to trigger a \newpage just before the given reference number to
% balance the columns on the last page. Adjust the reference number
% as needed - this may need to be readjusted if the document is 
% modified later.
% The "triggered" command can be changed if desired:
%\IEEEtriggercmd{\enlargethispage{-5in}}
%
% Alternatively, you can also directly use something like
% \enlargethispage{-7in}
% on the last page instead of breaking at a specific reference number.

% references section
%
% can use a bibliography generated by BibTeX as a .bbl file
% BibTeX documentation can be easily obtained at:
% http://mirror.ctan.org/biblio/bibtex/contrib/doc/
% The IEEEtran BibTeX style support page is at:
% http://www.michaelshell.org/tex/ieeetran/bibtex/
%\bibliographystyle{IEEEtran}
% argument is your BibTeX string definitions and bibliography database(s)
%\bibliography{IEEEabrv,../bib/paper}
%
% Before submitting to IEEE Sensors Letters, manually copy in the
% resultant .bbl file contents in place of the \bibliographystyle and
% \bibliography lines here:
\bibliographystyle{IEEEtran}
\bibliography{ref.bib}
% \begin{thebibliography}{1}

% \bibitem{IEEEhowto:kopka}
% H.~Kopka and P.~W. Daly, \emph{Guide to \LaTeX}, 4th~ed.\hskip 1em plus
%   0.5em minus 0.4em\relax Boston, MA: Addison-Wesley, 2004.

% \end{thebibliography}

% that's all folks
\end{document}